\documentclass[letterpaper]{article}
\usepackage{iccc}

\usepackage{amsmath}

\usepackage[pdftex]{graphicx}
\usepackage[font=footnotesize]{subfig}

\usepackage{times}
\usepackage{helvet}
\usepackage{courier}
\usepackage{array}

\title{Transforming Exploratory Creativity with DeLeNoX}
\author{Antonios Liapis$^1$, H\'{e}ctor P. Mart\'{i}nez$^2$, Julian Togelius$^1$ and Georgios N. Yannakakis$^2$\\
1: Center for Computer Games Research\\
IT University of Copenhagen\\
Copenhagen, Denmark\\
2: Institute of Digital Games\\
University of Malta\\
Msida, Malta\\
anli@itu.dk, hector.p.martinez@um.edu.mt, juto@itu.dk, georgios.yannakakis@um.edu.mt
}
\begin{document}
\maketitle 

\begin{abstract}
\begin{quote}
We introduce DeLeNoX (Deep Learning Novelty Explorer), a system that autonomously creates artifacts in constrained spaces according to its own evolving interestingness criterion. DeLeNoX proceeds in alternating phases of \emph{exploration} and \emph{transformation}. In the exploration phases, a version of novelty search augmented with constraint handling searches for maximally diverse artifacts using a given distance function. In the transformation phases, a deep learning autoencoder learns to compress the variation between the found artifacts into a lower-dimensional space. The newly trained encoder is then used as the basis for a new distance function, transforming the criteria for the next exploration phase. In the current paper, we apply DeLeNoX to the creation of spaceships suitable for use in two-dimensional arcade-style computer games, a representative problem in procedural content generation in games. We also situate DeLeNoX in relation to the distinction between exploratory and transformational creativity, and in relation to Schmidhuber's theory of creativity through the drive for compression progress.
\end{quote}
\end{abstract}

\section{Introduction}\label{introduction}

Within computational creativity research, many systems have been designed that create artifacts automatically through search in a given space for predefined objectives, using evolutionary computation or some similar stochastic global search/optimization algorithm. Recently, the novelty search paradigm has aimed to abandon all objectives, and simply search the space for a set of artifacts that is as diverse as possible, i.e. for maximum novelty \cite{lehman2011noveltysearch}. However, no search is without biases. Depending on the problem, the search space often contains constraints that limit and bias the exploration, while the mapping from genotype space (in which the algorithm searches) and phenotype space (in which novelty is calculated) is often indirect, introducing further biases. The result is a limited and biased novelty search, an incomplete exploration of the given space.

But what if we could characterize the bias of the search process as it unfolds and counter it? If the way space is being searched is continuously transformed in response to detected bias, the resulting algorithm would more thoroughly search the space by cycling through or subsuming biases. In applications such as game content generation, it would be particularly useful to sample the highly constrained space of useful artifacts as thoroughly as possible in this way.

In this paper, we present the Deep Learning Novelty Explorer (DeLeNoX) system, which is an attempt to do exactly this. DeLeNoX combines phases of exploration through constrained novelty search with phases of transformation through deep learning autoencoders. The target application domain is the generation of  two-dimensional spaceships which can be used in space shooter games such as \emph{Galaga} (Namco 1981). Automatically generating visually diverse spaceships which however fulfill constraints on believability addresses the ``content creation'' bottleneck of many game titles. 
The spaceships are generated by pattern-producing networks (CPPNs) via augmenting topologies \cite{stanley2006cppn}. In the exploration phases, DeLeNoX finds the most diverse set of spaceships possible given a particular distance function. In the transformation phases, it characterizes the found artifacts by obtaining a low-dimensional representation of their differences. This is done via autoencoders, a novel technique for nonlinear principal component analysis \cite{bengio2009deep}. The features found by the autoencoder are orthogonal to the bias of the current CPPN complexity, ensuring that each exploratory phase has a different bias than the previous. These features are then used to derive a new distance function which drives the next exploration phase. By using constrained novelty search for features tailored to the concurrent complexity, DeLeNoX can create content that is both \emph{useful} (as it lies within constraints) and \emph{novel}.

We will discuss the technical details of DeLeNoX shortly, and show results indicating that a surprising variety of spaceships can be found given the highly constrained search space. But first we will discuss the system and the core idea in terms of exploratory and transformational creativity, and in the context of Schmidhuber's theory of creativity as an impulse to improve the compressibility of growing data.

\section{Between exploratory and \\transformational creativity}

A ubiquitous distinction in creativity theory is that between exploratory and transformational creativity. Perhaps the most well-known statement of this distinction is due to Boden \shortcite{boden1990creative} and was later formalized by Wiggins \shortcite{wiggins2006framework} and others. However, similar ideas seem to be present in almost every major discussion of creativity such as ``thinking outside the box'' \cite{debono1970lateral}, ``paradigm shifts'' \cite{kuhn1962revolutions} etc. The idea requires that creativity is conceptualized as some sort of search in a space of artifacts or ideas. In Boden's formulation, exploratory creativity refers to search within a given search space, and transformational creativity refers to changing the rules that bind the search so that other spaces can be searched. Exploratory creativity is often associated with the kind of pedestrian problem solving that ordinary people engage in every day, whereas transformational creativity is associated with major breakthroughs that redefine the way we see problems. 

Naturally, much effort has been devoted to thinking up ways of modeling and implementing transformational creativity in a computational framework. Exploratory creativity is often modeled ``simply'' as objective-driven search, e.g. using constraint satisfaction techniques or evolutionary algorithms (including interactive evolution).

We see the distinction between exploratory and transformative creativity as a matter quantitative rather than qualitative. In some cases, exploratory creativity is indeed limited by hard \emph{constraints} that must be broken in order to transcend into unexplored regions of search space (and thus achieve transformational creativity). In other cases, exploratory creativity is instead limited by \emph{biases} in the search process. A painter might have a particular painting technique she defaults to, a writer a common set of plot devices he returns to, and an inventor might be accustomed to analyze problems in a particular order. This means that some artifacts are in practice never found, even though finding them would not break any constraints --- those artifacts are contained within the space delineated by the original constraints. Analogously, any search algorithm will over-explore some regions of search space and in practice never explore other areas because of particularities related to e.g. evaluation functions, variation operators or representation (cf. the discussion of search biases in machine learning~\cite{mitchell97learning}). This means that some artifacts are never found in practice, even though the representation is capable of expressing them and there exists a way in which they could in principle be found. 

\begin{figure}[tb]
\centering
\includegraphics[width=0.47\textwidth]{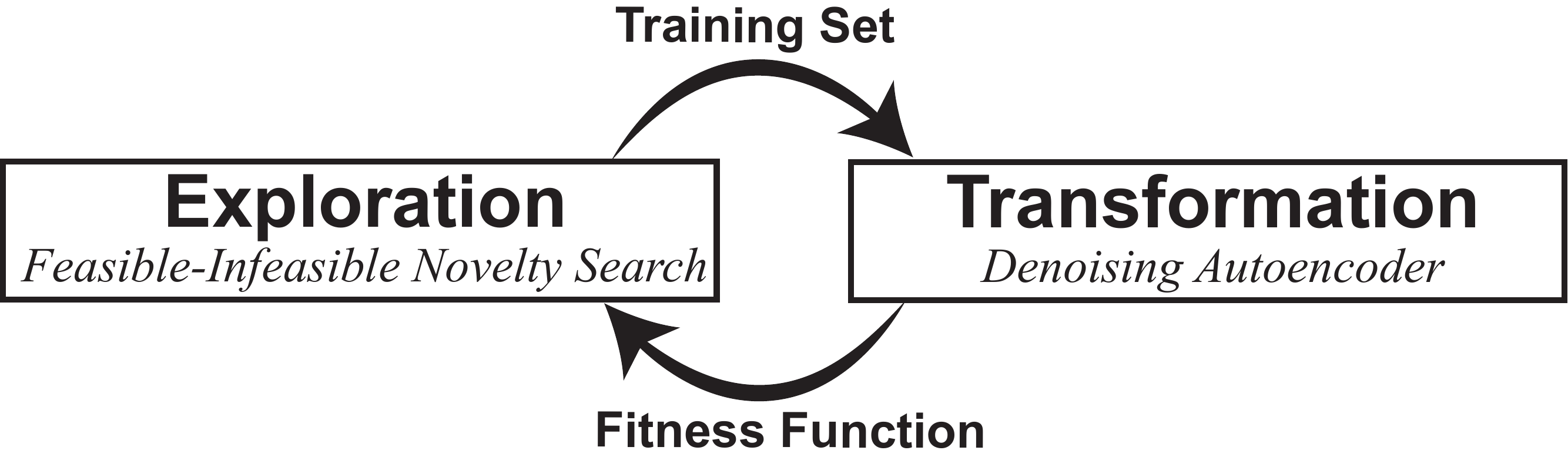}
\caption{Exploration transformed with DeLeNoX: the flowchart includes the general principles of DeLeNoX (bold) and the methods of the presented case study (italics).}
\label{fig:flowchart}
\end{figure}

\subsection{DeLeNoX and Transformed Exploration}

As mentioned above, the case study of this paper is two-dimensional spaceships. These are represented as images generated by Compositional Pattern-Producing Networks (CPPNs) with constraints on which shapes are viable spaceships. Exploration is done through a version of novelty search, which is a type of evolutionary algorithm that seeks to explore a search space as thoroughly as possible rather than maximizing an objective function. In order to do this, it needs a measure of difference between individuals. The distance measure inherently privileges some region of the search space over others, in particular when searching at the border of feasible search space. Additionally, CPPNs with different topologies are likely to create specific patterns in generated spaceships, with more complex CPPNs typically creating more complex patterns. Therefore, in different stages of this evolutionary complexification process, different regions of the search space will be under-explored. Many artifacts that are expressible within the representation will thus most likely not be found; in other words, there are limitations to creativity because of search biases.

In order to alleviate this problem and achieve a fuller coverage of space, we algorithmically \emph{characterize} the biases from the search process and the representation. This is what the autoencoders do. These autoencoders are applied on a set of spaceships resulting from an initial exploration of the space. A trained autoencoder is a function from a complete spaceship (phenotype) to a relatively low-dimensional array of real values. We then use the output of this function to compute a new distance measure, which differs from previous ones in that it better captures typical patterns at the current representational power of the spaceship-generating CPPNs. Changing the distance function amounts to changing the exploration process of novelty search, as novelty search is now in effect searching along different dimensions (see Fig.~\ref{fig:flowchart}). We have thus \emph{transformed} exploratory creativity, not by changing or abandoning any constraints, but by adjusting the search bias. This can be seen as analogous to changing the painting technique of a painter, the analysis sequence of an inventor, or introducing new plot devices for a writer. All of the spaceships that are found by the new search process could in principle have been found by the previous processes, but were very unlikely to be.

\section{Schmidhuber's theory of creativity}

Schmidhuber \shortcite{schmidhuber2006developmental,schmidhuber2007simple} advances an ambitious and influential theory of beauty, interestingness and creativity that arguably holds explanatory power at least under certain circumstances. Though the theory is couched in computational terms, it is meant to be applicable to humans and other animals as well as artificial agents. In Schmidhuber's theory, a beautiful pattern for a curious agent A is one that can successfully be compressed to much smaller description length by that agent's compression algorithm. However, perfect beauty is not interesting; an agent gets bored by environments it can compress very well and cannot learn to compress better, and also by those it cannot compress at all. Interesting environments for A are those which A can compress to some extent but where there is potential to improve the compression ratio, or in other words potential for A to learn about this type of environment. This can be illustrated by tastes in reading: beginning readers like to read linguistically and thematically simple texts, but such texts are seen by advanced readers as ``predictable'' (i.e. compressible), and the curious advanced readers therefore seek out more complex texts. In Schmidhuber's framework, creative individuals such as artists and scientists are also seen as a curious agents: they seek to pose themselves problems that are on the verge of what they can solve, learning as much as possible in the process. It is interesting to note the close links between this idea and the theory of flow \cite{csikszentmihalyi96creativity} but also theories of learning in children \cite{vygotsky1987collected} and game-players \cite{koster2004theory}. 

The DeLeNoX system fits very well into Schmidhuber's framework and can be seen as a novel implementation of a creative agent. The system proceeds in phases of exploration, carried out by novelty search which searches for interesting spaceships, and transformation, where autoencoders learn to compress the spaceships found in the previous exploration phase (see Fig.~\ref{fig:flowchart}) into a lower-dimensional representation. In the exploration phases, ``interesting'' amounts to far away from existing solutions according to the distance function defined by the autoencoder in the previous transformation phase. This corresponds to Schmidhuber's definition of interesting environments as those where the agent can learn (improve its compression for the new environment); the more distant the spaceships are, the more they force the autoencoder to change its compression algorithm (the weights of the network) in the next transformation phase. In the transformation phase, the learning in the autoencoder directly implements the improvement in capacity to compress recent environments (``compression progress'') envisioned in Schmidhuber's theory.

There are two differences between our model and Schmidhuber's model of creativity, however. In Schmidhuber's model, the agent stores all observations indefinitely and always retrains its compressor on the whole history of previous observations. As DeLeNoX resets its archive of created artifacts in every exploration phase, it is a rather forgetful creator. A memory could be implemented by keeping an archive of artifacts found by novelty search in all previous exploration phases, but this would incur a high and constantly increasing computational cost. It could however be argued that the dependence of each phase on the previous represents an implicit, decaying memory. The other difference to Schmidhuber's mechanism is that novelty search always looks for the solution/artifact that is most different to those that have been found so far, rather than the one predicted to improve learning the most. Assuming that the autoencoder compresses \emph{relatively} better the more diverse the set of artifacts is, this difference vanishes; this assumption is likely to be true at least in the current application domain.

\section{A case study of DeLeNoX: \\Spaceship Generation}\label{methodology}

This paper presents a case study of DeLeNoX for the creation of spaceship sprites, where exploration is performed via constrained novelty search which ensures a believable appearance, while transformation is performed via a denoising autoencoder which finds typical features in the spaceships' current representation (see Fig.~\ref{fig:flowchart}). Search is performed via neuroevolution of augmenting topologies, which changes the representational power of the genotype and warrants the transformation of features which bias the search.

\begin{figure}[tb]
\centering
\subfloat[CPPN]{\includegraphics[height=3cm]{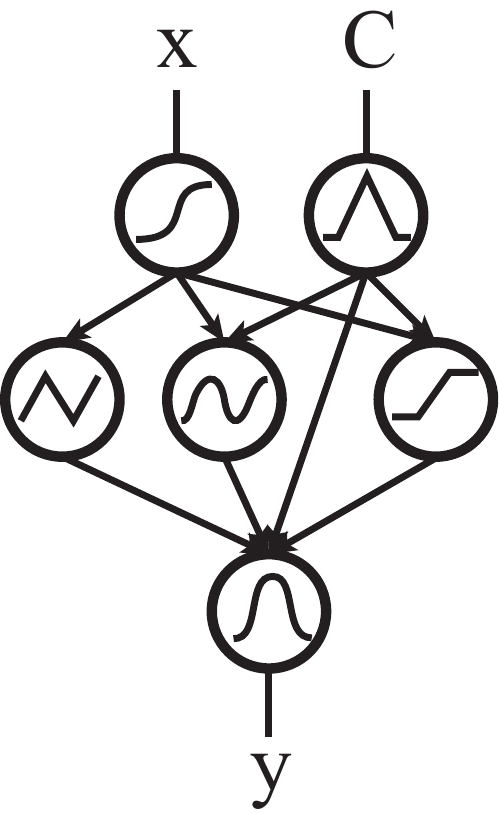}\label{fig:cppn}}\qquad
\subfloat[Spaceship representation]{\includegraphics[height=3cm]{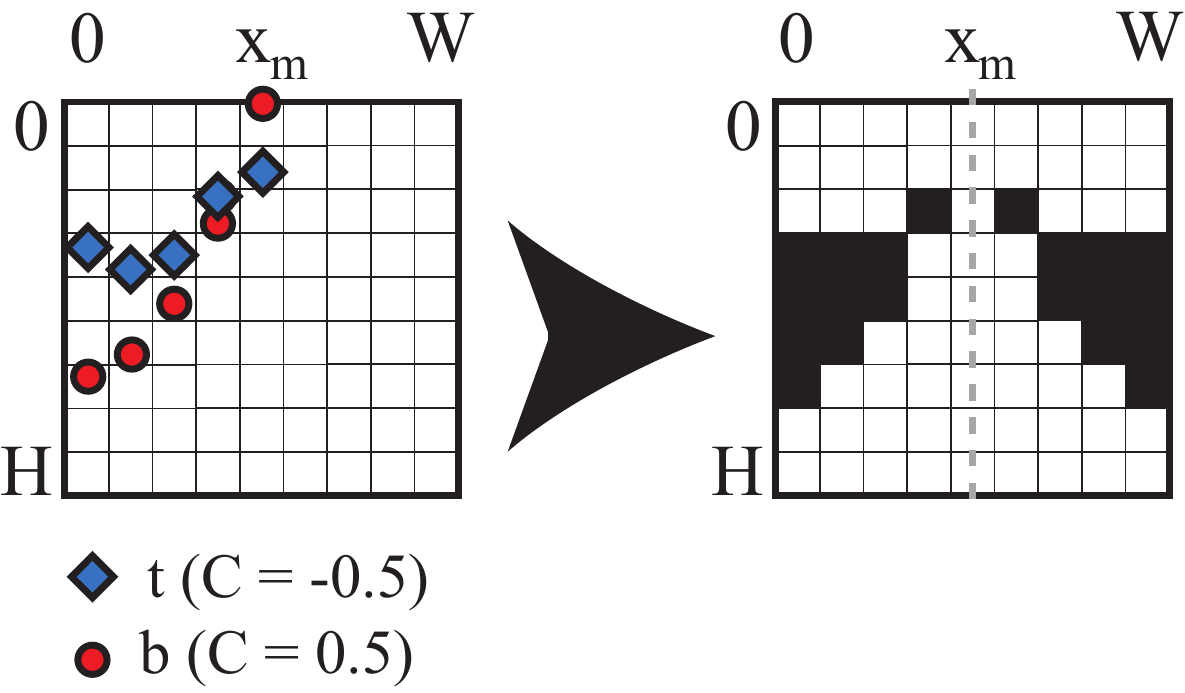}\label{fig:spaceship_generator}}
\caption{Fig~\ref{fig:cppn} shows a sample CPPN using the full range of pattern-producing activation functions available. Fig.~\ref{fig:spaceship_generator} shows the process of spaceship generation: the coordinates 0 to $x_m$, normalized as $0$ to $1$ (respectively) are used as input $x$ of the CPPN. Two $C$ values are used for each $x$, resulting in two points, top ($t$) and bottom ($b$) for each $x$. CPPN input $x$ and output $y$ are treated as the coordinates of $t$ and $b$;  if $t$ has a higher $y$ value than that of $b$ then the column is empty, else the hull extends between $t$ and $b$. The generated hull is reflected vertically along $x_m$.}
\label{fig:spaceship_methodology}
\end{figure}

\subsection{Domain Representation}\label{methodology_representation}

Spaceships are stored as two-dimensional sprites; the spaceship's hull is shown as black pixels. Each spaceship is encoded by a Compositional Pattern-Producing Network (CPPN), which is able to create complex patterns via function composition \cite{stanley2006cppn}. A CPPN is ideal for visual representation as it can be queried with arbitrary spatial granularity (infinite resolution); however, this study uses a fixed resolution for simplicity. Unlike standard artificial neural networks where all nodes have the same activation function, each CPPN node may have a different, pattern-producing function; six activation functions bound within $[0,1]$ are used in this study (see Fig.~\ref{fig:cppn}). To generate a spaceship, the sprite is divided into a number of equidistant columns equal to the sprite's width ($W$) in pixels. In each column, two points are identified as top ($t$) and bottom ($b$); the spaceship extends from $t$ to $b$, while no hull exists if $t$ is below $b$ (see Fig.~\ref{fig:spaceship_generator}). The $y$ coordinate of the top and bottom points is the output of the CPPN; its inputs are the point's $x$ coordinate and a constant $C$ which differentiates between $t$ and $b$ (with $C=-0.5$ and $C=0.5$, respectively). Only half of the sprites' columns, including the middle column at $x_m=\lceil\frac{W}{2}\rceil$, are used to generate $t$ and $b$; the remaining columns are derived by reflecting vertically along $x_m$.

A sufficiently expanded CPPN, as a superset of a multi-layer perceptron, is theoretically capable of representing any function. This means that any image could in principle be produced by a CPPN. However, the interpretation of CPPN output we use here means that images are severely limited to those where each column contains at most one vertical black bar. Additionally, the particularities of the NEAT complexification process, of the activation functions used and of the distance function which drives evolution make the system heavily biased towards particular shapes. It is this latter bias that is characterized within the transformation phase.

\subsection{Transformation Phase: Denoising Autoencoder}\label{methodology_autoencoder}

\begin{figure}[tb]
\centering
\includegraphics[width=0.45\textwidth]{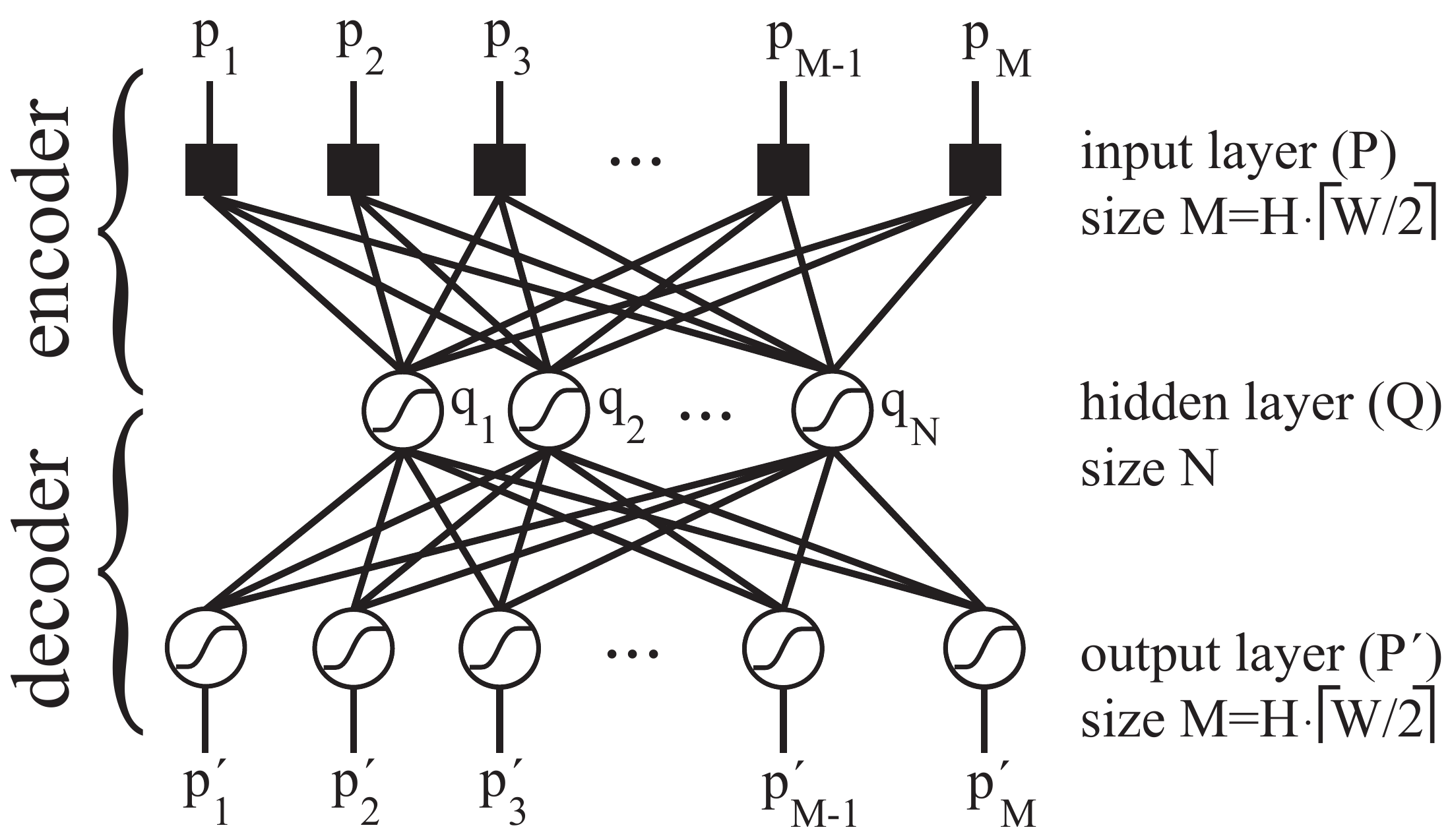}
\caption{The autoencoder architecture used for DeLeNoX, consisting of the encoder where $Q=f^\mathbf{w}(P)$ and the decoder where $P'=g^\mathbf{w}(Q)$. The higher-level representation in $q_1, q_2,\ldots,q_N$ is used to calculate the difference between individuals for the purposes of novelty search.}
\label{fig:autoencoder}
\end{figure}

The core innovation of DeLeNoX is the integration of autoencoders (AEs) in the calculation of the novelty heuristic (described in the next section), which is used to explore the search space according to the current representational power of the encoding CPPNs. 
AEs \cite{hinton1994autoencoders} are non-linear models that transform an input space $P$ into a new distributed representation $Q$ by applying a deterministic parametrized function called the \emph{encoder} $Q = f^{\mathbf{w}}(P)$. This encoder, instantiated in this paper as a single layer of logistic neurons, is trained alongside a \emph{decoder} (see Fig.~\ref{fig:autoencoder}) that maps back the transformed into the original representation ($P' = g^\mathbf{w}(Q)$) with a small reconstruction error, i.e. the original and corresponding decoded inputs are similar. By using a lower number of neurons than inputs, the AE is a method for the lossy compression of data; its most desirable feature, for the purposes of DeLeNoX, is that the compression is achieved by exploiting typical patterns observed in the training set.
In order to increase the robustness of this compression, we employ \emph{denoising autoencoders} (DAs), an AE variant that corrupts the inputs of the encoder during training  while enforcing that the original uncorrupted data is reconstructed \cite{vincent2008denoising}. Forced to both maintain most of the information from the input and undo the effect of corruption, the DA must ``capture the main variations in the data, i.e. on the manifold'' \cite{vincent2008denoising}, which makes DAs far more powerful tools than linear models for principal component analysis.

\begin{figure}[tb]
\centering
\subfloat[]{\includegraphics[height=1.8cm]{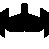}\label{fig:sampleShip3}}\hfill
\subfloat[]{\includegraphics[height=1.8cm]{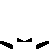}\label{fig:sampleShip1}}\hfill
\subfloat[]{\includegraphics[height=1.8cm]{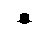}\label{fig:sampleShip2}}\hfill
\subfloat[]{\includegraphics[height=1.8cm]{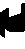}\label{fig:sampleShip4}}\hfill
\subfloat[]{\includegraphics[height=1.8cm]{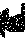}\label{fig:sampleShip5}}\hfill
\caption{Sample spaceships of 49 by 49 pixels, used for demonstrating DeLeNoX. Fig.~\ref{fig:sampleShip3} is a feasible spaceship; Fig.~\ref{fig:sampleShip1} and \ref{fig:sampleShip2} are infeasible, as they have disconnected pixels and insufficient size respectively. The autoencoder is trained to predict the left half of the spaceship in Fig.~\ref{fig:sampleShip3} (Fig.~\ref{fig:sampleShip4}) from a corrupted version of it (Fig.~\ref{fig:sampleShip5}).}
\label{fig:sampleShips}
\end{figure}
For the purposes of detecting the core visual features of the generated spaceships, DeLeNoX uses DAs to transform the spaceship's sprite to a low-dimensional array of real values, which correspond to the output of the encoder. Since spaceships are symmetrical along $x_m$, the training set consists of the left half of every spaceship sprite (see Fig.~\ref{fig:sampleShip4}). The encoder has $H{\cdot}{\lceil}\frac{W}{2}{\rceil}$ inputs ($P$), which are assigned a corrupted version of the spaceship's half-sprite; corruption is accomplished by randomly replacing pixels with 0, which is the same as randomly removing pixels from the spaceship (see Fig.~\ref{fig:sampleShip5}). The encoder has $N$ neurons, corresponding to the number of high-level features captured; each feature $q_i$ is a function of the input $P$ as $sig(W_i{\cdot}P+b_i)$ where $sig(x)$ the sigmoid function and $\{W_i,b_i\}$ the feature's learnable parameters (weight set and bias value, respectively). The output $P'$ of the decoder is an estimation of the uncorrupted half-sprite derived from $Q=[q_1, q_2, \ldots, q_N]$ via $P'=sig(W'{\cdot}Q+B')$; in this paper the DA uses tied weights and thus $W'$ is the transpose of $W=[W_1, W_2, \ldots, W_N]$. The parameters $\{W,B,B'\}$ are trained via backpropagation \cite{rumelhart1995backpropagation} according to the mean squared error between pixels in the uncorrupted half-sprite with those in the reconstructed sprite.

\subsection{Exploration Phase: Constrained Novelty Search}\label{methodology_evolution}

The spaceships generated by DeLeNoX are expected to be useful for a computer game; spaceships must have a believable appearance and sufficient size to be visible. Specifically, spaceships must not have disconnected pixels and must occupy at least half of the sprite's height and width; see examples of infeasible spaceships in Fig.~\ref{fig:sampleShip1} and \ref{fig:sampleShip2}. In order to optimize feasible spaceships towards novelty, content is evolved via a \emph{feasible-infeasible novelty search} (FINS) \cite{liapis2013gecco}. FINS follows the paradigm of the feasible-infeasible two-population genetic algorithm \cite{kimbrough2008fi2pop} by maintaining two separate populations: a \emph{feasible population} of individuals satisfying all constraints and an \emph{infeasible population} of individuals failing one or more constraints. Each population selects individuals among its own members, but feasible offspring of infeasible parents are transferred to the feasible population and vice versa; this form of interbreeding increases the diversity of both populations. In FINS, the feasible population selects parents based on a novelty heuristic ($\rho$) while the infeasible population selects parents based on their proximity to the feasible border ($f_{inf}$), defined as:
\begin{equation*}
f_{inf} =1-\frac{1}{3}\left[\max\{0,1-\tfrac{2w}{W}\}+\max\{0,1-\tfrac{2h}{H}\}+\tfrac{A_s}{A}\right]
\label{type:infeasible}
\end{equation*}
\noindent where $w$ and $h$ is the width and height of the spaceship in pixels; $W$ and $H$ is the width and height of the sprite in pixels; $A$ is the total number of black pixels on the image and $A_s$ the number of pixels on all disconnected segments. 

For the feasible population, the paradigm of novelty search is followed in order to explore the full spectrum of the CPPNs' representational power. The fitness score $\rho(i)$ for a feasible individual $i$ amounts to its average difference with the $k$ closest feasible neighbors within the population or in an archive of past novel individuals \cite{lehman2011noveltysearch}. In each generation, the $l$ highest-scoring feasible individuals are inserted in an archive of novel individuals. In DeLeNoX, the difference used to calculate $\rho$ is the Euclidean distance between the high-level features discovered by the denoising autoencoder; thus $\rho(i)$ is calculated as:
\begin{equation*}
\rho(i)=\frac{1}{k}\sum_{m=1}^{k}\sqrt{\sum_{n=1}^{N}\left[q_n(i)-q_n(\mu_m)\right]^2}
\label{eq:novelty_fitness}
\end{equation*}
where $\mu_m$ is the $m$-th-nearest neighbor of $i$ (in the population or the archive of novel individuals); $N$ is the number of hidden nodes (features) of the autoencoder and $q_n(i)$ the value of feature $n$ for spaceship $i$. As with the training process of the denoising autoencoder, the left half of spaceship $i$ is used as input to $q_n(i)$, although the input is not corrupted.

In both populations, evolution is carried out via neuroevolution of augmenting topologies \cite{stanley2002neat} using only mutation; an individual in the population may be selected (via fitness-proportionate roulette wheel selection) more than once for mutation. Mutation may add a hidden node (5\% chance), add a link (10\% chance), change the activation function of an existing node (5\% chance) or modify all links' weights by a small value.

\subsection{Experimentation}\label{experiment}

DeLeNoX will be demonstrated with the iteratively transformed exploration of spaceships on sprites of 49 by 49 pixels. The experiment consists of a series of \emph{iterations}, with each iteration divided into an exploration phase and a transformation phase. The exploration phase uses constrained novelty search to optimize a set of diverse spaceships, with ``diversity'' evaluated according to the features of the previous iteration; the transformation phase uses the set of spaceships optimized in the exploration phase to create new features which are better able to exploit the regularities of the current spaceship complexity. Each exploration phase creates a set of 1000 spaceships, which are generated from 100 independent runs of the FINS algorithm for 50 generations; the 10 fittest feasible individuals of each run are inserted into the set. Given the genetic operators used in the mutation scheme, each exploration phase augments the CPPN topology by roughly 5 nodes. While  the first iteration starts with an initial population consisting of CPPNs with no hidden nodes, subsequent iterations start with an initial population of CPPNs of the same complexity as the final individuals of the previous iteration. The total population of each run is 200 individuals, and parameters of novelty search are $k=20$ and $l=5$. Each evolutionary run maintains its own archive of novel individuals; no information regarding novelty is shared from previous iterations or across runs. Forgetting past visited areas of the search space is likely to hinder novelty search, but using a large archive of past individuals comes with a huge computational burden; given that CPPN topology augments in each iteration, it is less likely that previous novel individuals will be re-discovered, which makes  ``forgetting'' past breakthroughs an acceptable sacrifice.

Each transformation phase trains a denoising autoencoder with a hidden layer of 64 nodes, thus creating 64 high-level features. The weights and biases for these features are trained in the 1000 spaceships created in the exploration phase. Training runs for 1000 epochs, trying to accurately predict the real half-sprite of the spaceship (see Fig.~\ref{fig:sampleShip4}) from a corrupted version of it (see Fig.~\ref{fig:sampleShip5}); corruption occurs by replacing any pixel with a white pixel (with 10\% chance).

We observe the progress of DeLeNoX for 6 iterations. For the first iteration, the features driving the exploration phase are trained on a set of 1000 spaceships created by randomly initialized CPPNs with no hidden nodes; these spaceships and features are identified as ``initial''. 
The impact of transformation is shown via a second experiment, where spaceships evolve for 6 iterations using the initial set of features trained from simple spaceships with no transformation phases between iterations; this second experiment is named ``static'' (contrary to  the proposed ``transforming'' method). 

\begin{figure}[tb]
\centering
\small
$\begin{array}{ c || c | c | c | c | c | c | c}
$Iter.$ & $Initial$ & $1st$ & $2nd$ & $3rd$ & $4th$ & $5th$ & $6th$\\
\hline\hline
$Best$ &
\subfloat{\includegraphics[width=0.65cm]{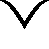}}&
\subfloat{\includegraphics[width=0.65cm]{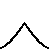}}&
\subfloat{\includegraphics[width=0.65cm]{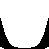}}&
\subfloat{\includegraphics[width=0.65cm]{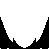}}&
\subfloat{\includegraphics[width=0.65cm]{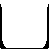}}&
\subfloat{\includegraphics[width=0.65cm]{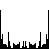}}&
\subfloat{\includegraphics[width=0.65cm]{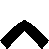}}
\\
&
\subfloat{\includegraphics[width=0.65cm]{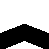}}&
\subfloat{\includegraphics[width=0.65cm]{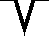}}&
\subfloat{\includegraphics[width=0.65cm]{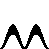}}&
\subfloat{\includegraphics[width=0.65cm]{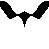}}&
\subfloat{\includegraphics[width=0.65cm]{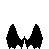}}&
\subfloat{\includegraphics[width=0.65cm]{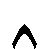}}&
\subfloat{\includegraphics[width=0.65cm]{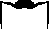}}
\\
&
\subfloat{\includegraphics[width=0.65cm]{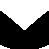}}&
\subfloat{\includegraphics[width=0.65cm]{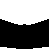}}&
\subfloat{\includegraphics[width=0.65cm]{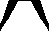}}&
\subfloat{\includegraphics[width=0.65cm]{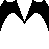}}&
\subfloat{\includegraphics[width=0.65cm]{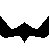}}&
\subfloat{\includegraphics[width=0.65cm]{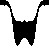}}&
\subfloat{\includegraphics[width=0.65cm]{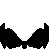}}
\\
&
\subfloat{\includegraphics[width=0.65cm]{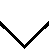}}&
\subfloat{\includegraphics[width=0.65cm]{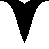}}&
\subfloat{\includegraphics[width=0.65cm]{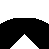}}&
\subfloat{\includegraphics[width=0.65cm]{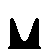}}&
\subfloat{\includegraphics[width=0.65cm]{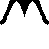}}&
\subfloat{\includegraphics[width=0.65cm]{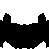}}&
\subfloat{\includegraphics[width=0.65cm]{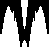}}
\\
&
\subfloat{\includegraphics[width=0.65cm]{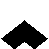}}&
\subfloat{\includegraphics[width=0.65cm]{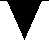}}&
\subfloat{\includegraphics[width=0.65cm]{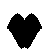}}&
\subfloat{\includegraphics[width=0.65cm]{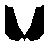}}&
\subfloat{\includegraphics[width=0.65cm]{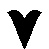}}&
\subfloat{\includegraphics[width=0.65cm]{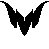}}&
\subfloat{\includegraphics[width=0.65cm]{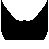}}
\\ 
$Worst$ &
\subfloat{\includegraphics[width=0.65cm]{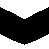}}&
\subfloat{\includegraphics[width=0.65cm]{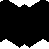}}&
\subfloat{\includegraphics[width=0.65cm]{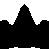}}&
\subfloat{\includegraphics[width=0.65cm]{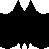}}&
\subfloat{\includegraphics[width=0.65cm]{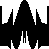}}&
\subfloat{\includegraphics[width=0.65cm]{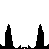}}&
\subfloat{\includegraphics[width=0.65cm]{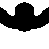}}
\end{array}$
\caption{Sample spaceships among the results of each iteration of exploration; such spaceships comprise the training set for detecting the next iteration's features (transforming run). The best and worst spaceship in terms of difference (using the previous iteration's features) is included, along with spaceships evenly distributed in terms of difference.}
\label{fig:samplespaceships}
\end{figure}

\begin{figure}[tb]
\centering
$\begin{array}{ c || c | c | c | c | c | c | c}
$Iter.$ & $Initial$ & $1st$ & $2nd$ & $3rd$ & $4th$ & $5th$ & $6th$ \\
\hline\hline
$Best$ &
\subfloat{\includegraphics[width=0.65cm]{./graphics/results/initial/ship0.png}}&
\subfloat{\includegraphics[width=0.65cm]{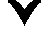}}&
\subfloat{\includegraphics[width=0.65cm]{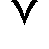}}&
\subfloat{\includegraphics[width=0.65cm]{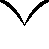}}&
\subfloat{\includegraphics[width=0.65cm]{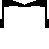}}&
\subfloat{\includegraphics[width=0.65cm]{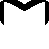}}&
\subfloat{\includegraphics[width=0.65cm]{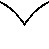}}
\\
&
\subfloat{\includegraphics[width=0.65cm]{./graphics/results/initial/ship200.png}}&
\subfloat{\includegraphics[width=0.65cm]{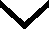}}&
\subfloat{\includegraphics[width=0.65cm]{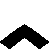}}&
\subfloat{\includegraphics[width=0.65cm]{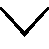}}&
\subfloat{\includegraphics[width=0.65cm]{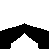}}&
\subfloat{\includegraphics[width=0.65cm]{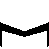}}&
\subfloat{\includegraphics[width=0.65cm]{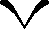}}
\\
&
\subfloat{\includegraphics[width=0.65cm]{./graphics/results/initial/ship400.png}}&
\subfloat{\includegraphics[width=0.65cm]{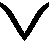}}&
\subfloat{\includegraphics[width=0.65cm]{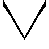}}&
\subfloat{\includegraphics[width=0.65cm]{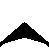}}&
\subfloat{\includegraphics[width=0.65cm]{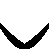}}&
\subfloat{\includegraphics[width=0.65cm]{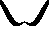}}&
\subfloat{\includegraphics[width=0.65cm]{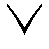}}
\\
&
\subfloat{\includegraphics[width=0.65cm]{./graphics/results/initial/ship600.png}}&
\subfloat{\includegraphics[width=0.65cm]{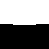}}&
\subfloat{\includegraphics[width=0.65cm]{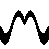}}&
\subfloat{\includegraphics[width=0.65cm]{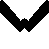}}&
\subfloat{\includegraphics[width=0.65cm]{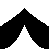}}&
\subfloat{\includegraphics[width=0.65cm]{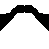}}&
\subfloat{\includegraphics[width=0.65cm]{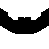}}
\\
&
\subfloat{\includegraphics[width=0.65cm]{./graphics/results/initial/ship800.png}}&
\subfloat{\includegraphics[width=0.65cm]{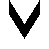}}&
\subfloat{\includegraphics[width=0.65cm]{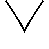}}&
\subfloat{\includegraphics[width=0.65cm]{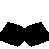}}&
\subfloat{\includegraphics[width=0.65cm]{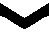}}&
\subfloat{\includegraphics[width=0.65cm]{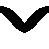}}&
\subfloat{\includegraphics[width=0.65cm]{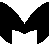}}
\\
$Worst$ &
\subfloat{\includegraphics[width=0.65cm]{./graphics/results/initial/ship999.png}}&
\subfloat{\includegraphics[width=0.65cm]{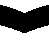}}&
\subfloat{\includegraphics[width=0.65cm]{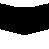}}&
\subfloat{\includegraphics[width=0.65cm]{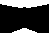}}&
\subfloat{\includegraphics[width=0.65cm]{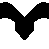}}&
\subfloat{\includegraphics[width=0.65cm]{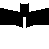}}&
\subfloat{\includegraphics[width=0.65cm]{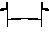}}
\end{array}$
\caption{Sample spaceships (sorted by difference) among the results of each iteration of exploration driven by static features trained on the initial spaceship set (static run).}
\label{fig:samplespaceshipsStatic}
\end{figure}

The final spaceships generated in the exploration phase of each iteration are shown in Fig.~\ref{fig:samplespaceships} for the transforming run and in Fig.~\ref{fig:samplespaceshipsStatic} for the static run. For the purposes of brevity, the figures show six samples selected based on their diversity (according to the features on which they were evolved);  Fig.~\ref{fig:samplespaceships} and \ref{fig:samplespaceshipsStatic} therefore not only showcase the artifacts generated by DeLeNoX, but the sampling method demonstrates 
the shapes which are identified as ``different'' by the features.

In Fig.~\ref{fig:samplespaceships}, the shifting representational power of CPPNs is obvious: CPPNs with no hidden nodes tend to create predominantly V-shaped spaceships, while larger networks create more curved shapes (such as in the 2nd iteration) and eventually lead to jagged edges or ``spikes'' in later iterations. While CPPNs can create more elaborate shapes with larger topologies, Fig.~\ref{fig:samplespaceships} includes simple shapes even in late iterations: such an example is the 6th iteration, where two of the sample spaceships seem simple. This is likely due to the lack of a ``long-term memory'', since there is no persistent archive of novel individuals across iterations.

\begin{figure*}
\centering
\subfloat[Initial]{\includegraphics[width=0.125\textwidth]{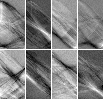}\label{fig:initialfilters}}\hfill
\subfloat[1st]{\includegraphics[width=0.125\textwidth]{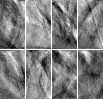}}\hfill
\subfloat[2nd]{\includegraphics[width=0.125\textwidth]{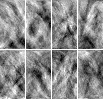}}\hfill
\subfloat[3rd]{\includegraphics[width=0.125\textwidth]{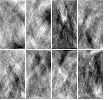}}\hfill
\subfloat[4th]{\includegraphics[width=0.125\textwidth]{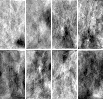}}\hfill
\subfloat[5th]{\includegraphics[width=0.125\textwidth]{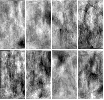}}\hfill
\subfloat[6th]{\includegraphics[width=0.125\textwidth]{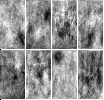}}\hfill
\caption{A sample of the 64 trained features at the end of each iteration. The visualization displays the weights of each pixel of the input (i.e. the left half of the spaceship's sprite). Weights are normalized to black (lowest) and white (highest).}
\label{fig:samplefilters}
\end{figure*}

In terms of detected features, Fig.~\ref{fig:samplefilters} displays a random sample of the 64 features trained in each transformation phase of the transforming run; the static run uses the ``initial'' features (see Fig.~\ref{fig:initialfilters}) in every iteration. The shape of the spaceships directly affects the features' appearance: for instance, the simple V-shaped spaceships of the initial training set result in features which detect diagonal edges. The features become increasingly more complex, and thus difficult to identify, in later iterations: while in the 1st iteration straight edges are still prevalent, features in the 5th or 6th iterations detect circular or vertical areas.

Comparing Fig.~\ref{fig:samplespaceshipsStatic} with Fig.~\ref{fig:samplespaceships}, we observe that despite the larger CPPN topologies of later iterations, spaceships evolved in the static run are much simpler than their respective ones in the transforming run. Exploration in the static run is always driven by simple initial features (see Fig.~\ref{fig:initialfilters}), showing how the features used in the fitness function $\rho$ bias search. On the contrary, the transformation phase in each iteration counters this bias and re-aligns exploration towards more visually diverse artifacts.

\renewcommand{\arraystretch}{0.8}% Tighter

\begin{figure}[tb]
\centering
\subfloat{\includegraphics[width=40mm, trim = 40mm 14mm 30mm 23mm, clip]{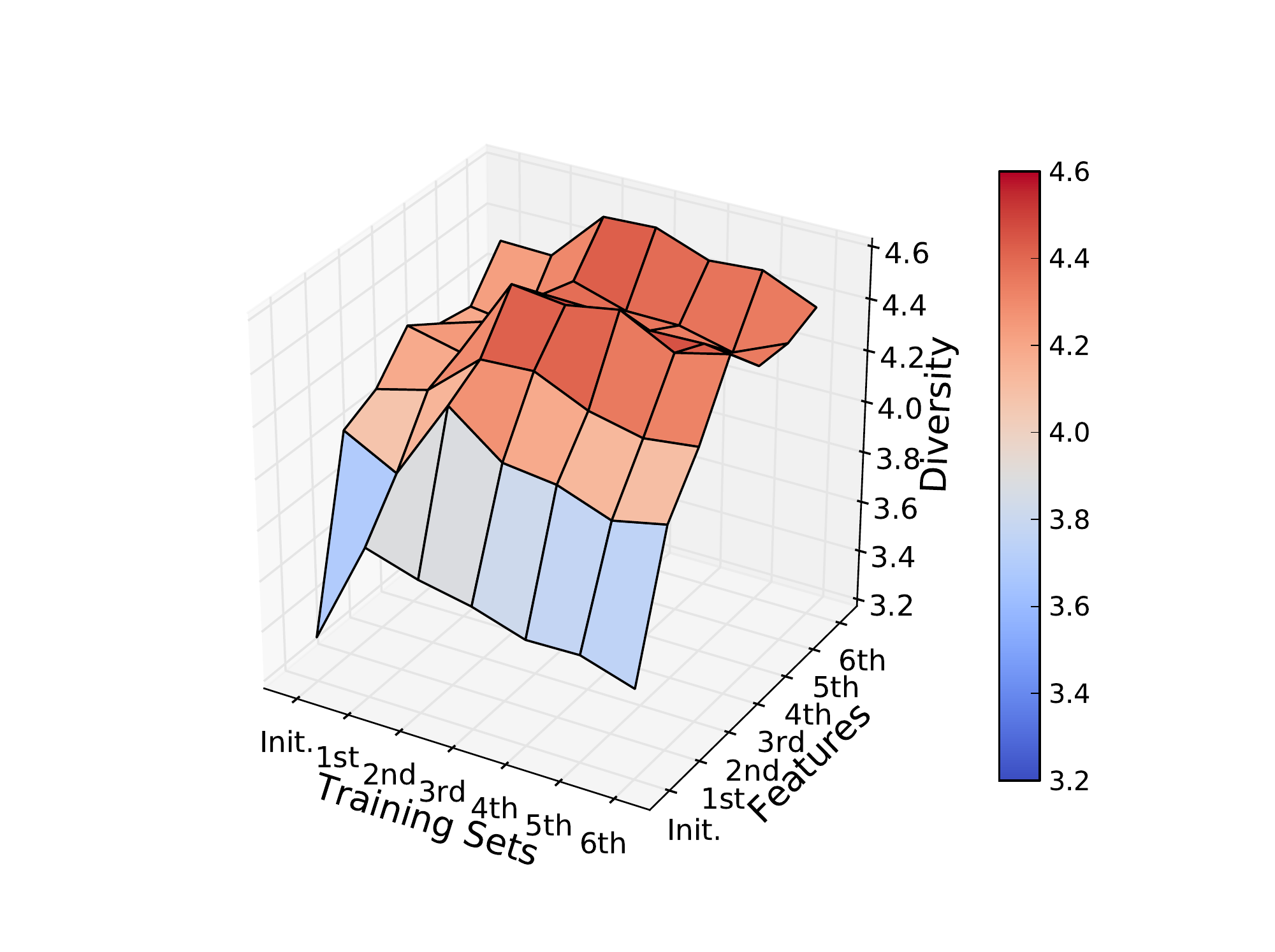}} \hfill
\subfloat{\includegraphics[width=40mm, trim = 40mm 14mm 30mm 23mm, clip]{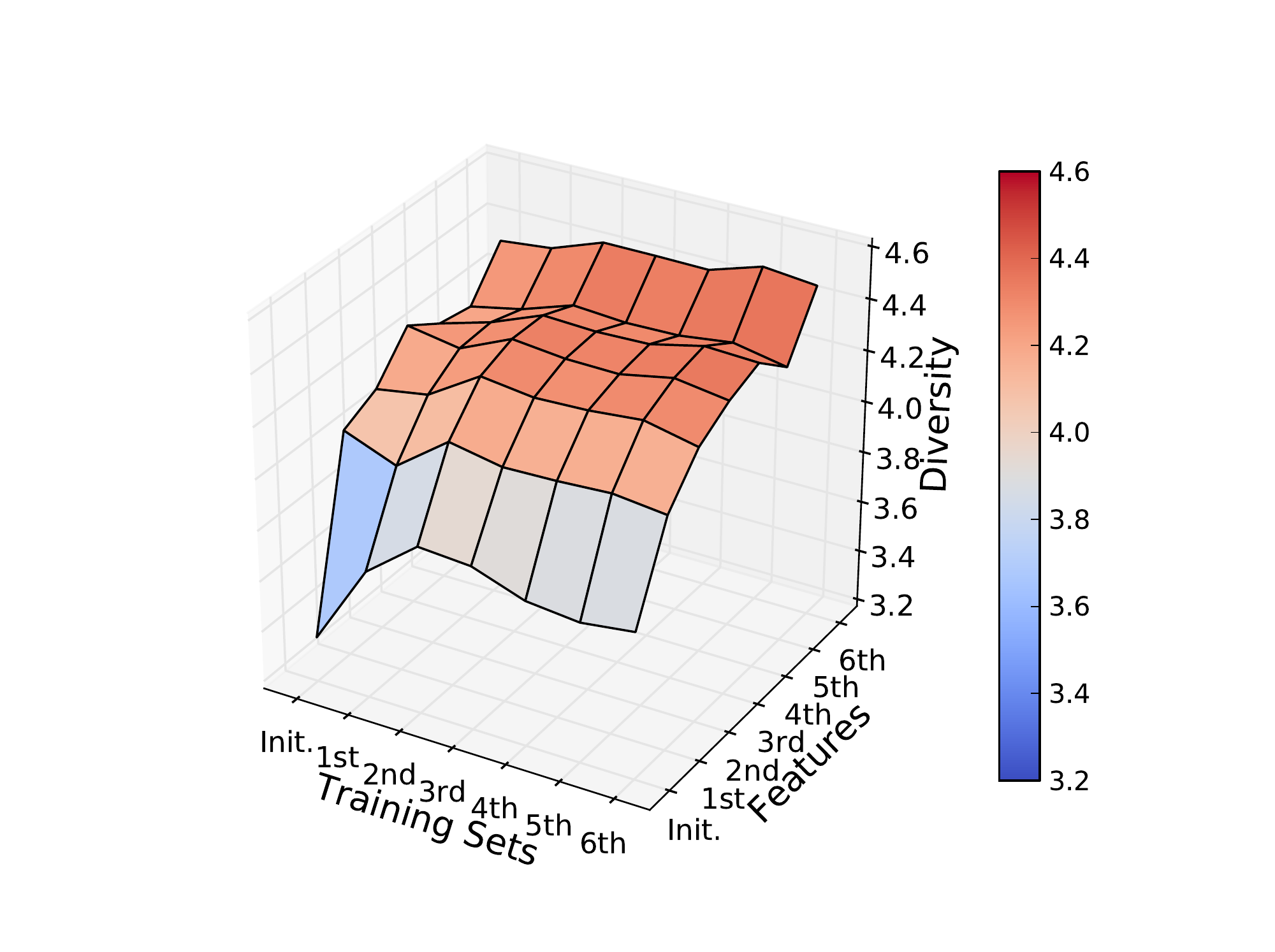}}\hfill
\caption{Diversity scores of the training sets at the end of each iteration's exploration phase, derived from the feature sets trained in the transformation phases of the transforming run. The training sets of the transforming run are evaluated on the left figure, and those of the static run on the right.}
\label{fig:averagedifference}
\end{figure}

The diversity of spaceships and the quality of detected features can be gleaned from Fig.~\ref{fig:averagedifference}, in which features trained in different iterations of the transforming run generate distance metrics which evaluate the diversity of every iteration's training set, both for the transforming and for the static run. Diversity is measured as the Euclidean distance averaged from all spaceship pairs of the training set of an iteration. In the transforming run, the highest diversity score for a feature set is usually attained in the training set of the following iteration (e.g. the initial features score the highest diversity in the 1st iteration's spaceships). This is expected, since the features of the previous iteration are used in the distance function driving novelty search in the next iteration. This trend, however, does not hold in the last 3 iterations, possibly because patterns after the 3rd iteration become too complex for 64 features to capture, while the simpler patterns of earlier iterations are more in tune with what they can detect. It is surprising that features of later iterations, primarily those of the 3rd and 6th iteration, result in high diversity values in most training sets, even those of the static run which were driven by the much simpler initial features. It appears that features trained in the more complicated shapes of later iterations are more general --- as they can detect patterns they haven't actually seen, such as those in the static run --- than features of the initial or 1st iteration which primarily detect straight edges (see Fig.~\ref{fig:samplefilters}). 

\section{Discussion}\label{discussion}

This paper has presented DeLeNoX as a system which transforms exploration of the search space in order to counter the biases of the representation and the evolutionary process. While short, the included case study demonstrates the potential of DeLeNoX in several distinct but complementary ways. The shifting representation of augmenting CPPNs benefits from the iterative transformations of the novelty heuristic which is used to evolve it, as demonstrated by early features which detect straight lines versus later features which focus on areas of interest. Using the early, simple features for evolving complex CPPNs is shown to hinder exploration since the representational bias which caused those features to be prevalent has been countered by augmenting topologies. On the other hand, the iterative exploration guided by features tailored to the representation creates a more diverse training set for the autoencoder, resulting in an overall improvement in the features detected as shown by the increased diversity scores of later features on the same data. This positive feedback loop, where the exploration phase benefits from the transformation phase, which in turn benefits from the improved divergent search of exploration is the core argument for DeLeNoX. It should be noted, however, that for this case study DeLeNoX is not without its own biases, as the increasingly diverse training set eventually challenges the feature detector's ability to capture typical patterns in the latest of presented iterations; suggestions for countering such biases will be presented in this section.

The case study presented in this paper is an example of exploration via high-level features derived by compressing information based on their statistical dependencies. The number of features chosen was arguably arbitrary; it allows for a decent compression (980 pixels to 64 real values) and measuring the Euclidean distance for novelty search is computationally manageable. At the same time, it is large enough to capture the most prevalent features among generated spaceships, at least in the first iterations where spaceships and their encoding CPPNs are simple. As exploration becomes more thorough --- enhanced both by the increased representational power of larger CPPNs and by more informed feature detectors --- typical patterns become harder to find. It could be argued that as exploration results in increasingly more diverse content, the number of features should increase to counter the fewer dependencies in the training set; for the same reasons, the size of the training set should perhaps increase. Future experiments should evaluate the impact of the number of features and the size of the training set both on the accuracy of the autoencoder and on the progress of novelty search. Other experiments should explore the potential of adjusting these values dynamically on a per-iteration basis; adjustments can be made via a constant multiplier or according to the quality of generated artifacts.

It should be pointed out that the presented case study uses a single autoencoder, which is able to discover simple features such as edges. These simple features are easy to present visually, and deriving the distance metric is straightforward based on the outputs of the autoencoder's hidden layer. For a simple testbed such as spaceship generation, features discovered by the single autoencoder suffice --- especially in early iterations of novelty search. However, the true potential of DeLeNoX will be shown via stacked autoencoders which allow for truly \emph{deep} learning; the outputs from the upper layers of such a deep belief network \cite{bengio2009deep} represent more ``abstract'' concepts than those of a single autoencoder. Using such robust features for deriving a novelty value is likely to address current limitations of the feature extractor in images generated by complex CPPNs, and can be applied to more complex problems.

The case study presented in this paper is ideal for demonstrating DeLeNoX due to the evolutionary complexification of CPPNs; the indirect mapping between genotype and phenotype and the augmenting topologies both warrant the iterative transformation of the features which drive novelty search. A direct or static mapping would likely find the iterative transformation of the search process less useful, since representational bias remains constant. However, any indirect mapping between genotype and phenotype including neuroevolution, grammatical evolution or genetic programming can be used for DeLeNoX.

\section{Related Work}\label{literature}

DeLeNoX is indirectly linked to the foci of a few studies in automatic content generation and evolutionary art. The creation of artifacts has been the primary focus of evolutionary art; however, the autonomy of art generation is often challenged by the use of interactive evolution 
driven by human preferences. In order to create closed systems, an art appreciation component is used to automatically evaluate generated artifacts. This \emph{artificial art critic} \cite{machado2003critics} is often an artificial neural network pre-trained to simulate user ratings in a collection of generated content \cite{baluja1999automated} or between man-made and generated images \cite{machado2007artists}. Image compression has also been used in the evaluation of generated artifacts \cite{machado2007artists}. While DeLeNoX essentially uses an artificial neural network to learn features of the training set, it does not simulate human aesthetic criteria as its training is unsupervised; moreover, the learned features are used to diversify the generated artifacts rather than converge them towards a specific art style or aesthetic. This same independence from human aesthetics, however, makes evaluating results of DeLeNoX difficult. Finally, while the autoencoder compresses images to a much smaller size, this compression is tailored to the particularities of the training set, unlike the generic compression methods such as \emph{jpeg} used in NEvAr \cite{machado2007artists}.  Recent interest in dynamically extracting features targeting deviation from previously evolved content \cite{machado2013novelty} has several similarities to DeLeNoX; the former approach, however, does not use novelty search (and thus exploration of the search space is limited) while features are extracted via supervised learning on a classification task between newly (and previously) generated artifacts and man-made art pieces.

The potential of DeLeNoX is demonstrated using the generation of spaceship sprites as a testbed. Spaceship generation is representative of the larger problem of automatic game content creation which has recently received considerable academic interest \cite{yannakakis2012revisited}. Search-based techniques such as genetic algorithms are popular for optimizing many different properties of game content; for a full survey see \cite{togelius2011searchbased}.
Procedurally generated spaceships have been optimized, via neuroevolution, for performance measures such as speed \cite{liapis2011functional} or for predefined aesthetic measures such as symmetry \cite{liapis2012adaptivemodel,liapis2011visualproperties}. Similarly to the method described in this paper, these early attempts use CPPN-NEAT to generate a spaceship's hull. This paper, however, describes a spaceship via top and bottom points and uses a sprite-based representation, both of which are more likely to generate feasible content; additionally, the spaceship's thrusters and weapons are not considered.

\section{Acknowledgments}
The research is supported, in part, by the FP7 ICT project SIREN (project no: 258453) and by the FP7 ICT project C2Learn (project no: 318480).

\bibliographystyle{iccc}
\bibliography{DeLeNoX}

\end{document}